\documentclass[sigconf,nonacm=true]{acmart}
\usepackage{booktabs} 
\usepackage[ruled]{algorithm2e} 

\SetAlFnt{\small}
\SetAlCapFnt{\small}
\SetAlCapNameFnt{\small}
\SetAlCapHSkip{0pt}
\usepackage{multirow}

\newcommand{\hl}[1]{{#1}}

\begin{document}
\title{Fusing Monocular Images and Sparse IMU Signals for Real-time Human Motion Capture}

\author{Shaohua Pan}
\email{shaohua-pan@outlook.com}
\affiliation{%
  \institution{School of software and BNRist, Tsinghua University}
  \country{China}
}

\author{Qi Ma}
\email{mq19@mails.tsinghua.edu.cn}
\affiliation{%
  \institution{School of software and BNRist, Tsinghua University}
  \country{China}
}

\author{Xinyu Yi}
\email{yixy20@mails.tsinghua.edu.cn}
\affiliation{%
  \institution{School of software and BNRist, Tsinghua University}
  \country{China}
}

\author{Weifeng Hu}
\email{huweifeng@oppo.com}
\affiliation{%
  \institution{OPPO Research Institute}
  \country{China}
}

\author{Xiong Wang}
\email{wangxiong1@oppo.com}
\affiliation{%
  \institution{OPPO Research Institute}
  \country{China}
}

\author{Xingkang Zhou}
\email{zhouxingkang@oppo.com}
\affiliation{%
  \institution{OPPO Research Institute}
  \country{China}
}

\author{Jijunnan Li}
\email{lijijunnan@oppo.com}
\affiliation{%
  \institution{OPPO Research Institute}
  \country{China}
}

\author{Feng Xu}
\email{xufeng2003@gmail.com}
\affiliation{%
  \institution{School of software and BNRist, Tsinghua University}
  \country{China}
}

\begin{abstract}
    Either RGB images or inertial signals have been used for the task of motion capture (mocap), but combining them together is a new and interesting topic.
We believe that the combination is complementary and able to solve the inherent difficulties of using one modality input, including occlusions, extreme lighting/texture, and out-of-view for visual mocap and global drifts for inertial mocap.
To this end, we propose a method that fuses monocular images and sparse IMUs for real-time human motion capture.
Our method contains a dual coordinate strategy to fully explore the IMU signals with different goals in motion capture.
To be specific, besides one branch transforming the IMU signals to the camera coordinate system to combine with the image information, there is another branch to learn from the IMU signals in the body root coordinate system to better estimate body poses.
%
Furthermore, a hidden state feedback mechanism is proposed for both two branches to compensate for their own drawbacks in extreme input cases. 
Thus our method can easily switch between the two kinds of signals or combine them in different cases to achieve a robust mocap.
%
%
Quantitative and qualitative results demonstrate that by delicately designing the fusion method, our technique significantly outperforms the state-of-the-art vision, IMU, and combined methods on both global orientation and local pose estimation.
Our codes are available for research at \url{https://shaohua-pan.github.io/robustcap-page/}.
\end{abstract}

\begin{teaserfigure}
  \includegraphics[width=\textwidth, height=.27\linewidth]{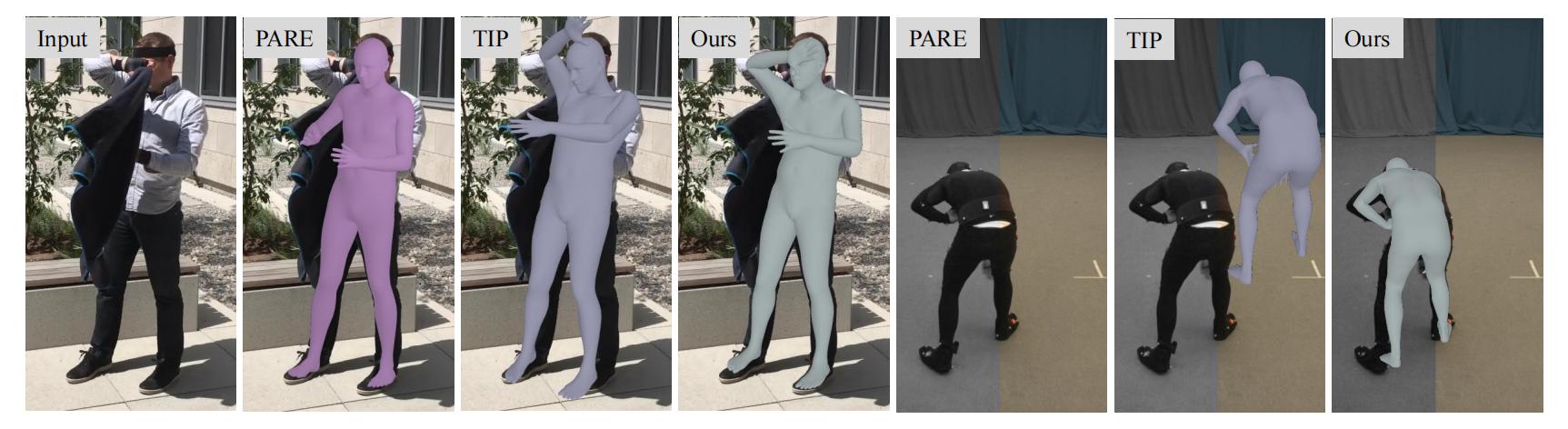}
  \caption{Comparison with state-of-the-art vision-based method PARE \cite{PARE} and IMU-based method TIP \cite{TIP}. Left (user in the camera view): PARE suffers occlusion and uniform texture, and TIP suffers pose ambiguity. Right (user out of the camera view, gray region indicating the outside): PARE cannot see the user, and TIP suffers the drift artifact (our method could use previous visible frames to reduce the drift). }
  \label{fig:teaser}
\end{teaserfigure}

\maketitle
\section{Introduction}
Human motion capture (mocap) is an important research topic in computer vision and graphics and has many applications in virtual reality, medical care, games, and animation.
Traditional multi-view techniques achieve high accuracy and robustness, but they are heavy and expensive and thus difficult to be used by end users.
Recently, with the development of mobile sensing techniques, RGB cameras and Inertial Measurement Units (IMUs) have been integrated into daily devices like phones, smartwatches, and eyeglasses. 
They can be easily accessed, and mocap with these sensors will have much broader applications.
%
%
%
%
%
\par
Human motion can be captured from a monocular camera or, alternatively, a sparse set of IMUs.
%
%
The former~\cite{ROMP,spin,PARE,VIBE} has become increasingly effective with the development of deep learning techniques.
However, they still fail in extreme lighting conditions, severe occlusions, or humans moving out of the camera view.
%
%
%
While IMU-based methods~\cite{DIP,TransPose,PIP,TIP} can get rid of these limitations, they cannot accurately estimate human translation due to the substantial drifts caused by sensor error accumulation.
%
%
%
\par
We propose to fuse monocular images with sparse IMUs to achieve real-time mocap with high accuracy and robustness for estimating both body pose and translation.
Our method fuses visual and inertial information to estimate better human motion when the performer is in the camera view.
While for some extreme cases where the performer is severely occluded, poorly lighted, or moving out of the camera view, our technique majorly uses IMUs to track the human motion.
Compared with the pure visual solutions, our technique outputs plausible mocap results robustly in the aforementioned extreme cases. 
On the other hand, when the performer is visible to the camera, the visual information is used as an online calibration to solve the drifting problem caused by cumulative errors in the inertial signals.
So the inertial and visual information leverage each other in our system to achieve a real-time lightweight mocap with high accuracy and robustness.
%
%
%
%
\par

Our technique is based on a dual coordinate strategy, which makes the model training adaptive to the two input modalities.
Specifically, the IMU signals are transformed into the camera coordinate system when integrated with visual signals but are processed in the human root coordinate system when used solely.
This design fully utilizes the characteristics of the two input modalities: visual signals capture both global position and local pose information of humans, while inertial measurements majorly track the human's local movements with global motion drift caused by inertial error accumulation. 
When the visual signal is available and confident, we align the IMU signal to the camera coordinate system as a complement, in which the human's absolute pose and motion dynamics are both well represented, and thus learning in this coordinate system makes full use of the visual signal.
%
%
Differently, when the inertial measurements are used solely (visual signals not available nor confident), the camera coordinate system becomes a suboptimal choice as the global part of the input modalities (\textit{i.e.}, visual signals) does not exist.
We thus seek a change from a global "third-person perspective" (\textit{i.e.}, camera coordinate system) to a local "first-person perspective" (\textit{i.e.}, root coordinate system) to capture the human motion.
The key is that human motions may be seen differently from the camera's view but are exactly the same from a local view, \textit{e.g.,} walking straight in different directions.
To this end, the problem is simplified, and we achieve better results in the root coordinated system for local pose estimation.
In summary, the delicate design of the dual coordinate strategy for the two branches explores the inertial signals as much as possible in different use cases.

\par
Within the dual coordinate strategy, hidden state feedback is further proposed to enable information exchange between the two coordinate systems.
In either coordinate system, human motion is estimated independently based on the historical input signals through deep temporal modeling.
As a result, the internal state in the coordinate system may deviate from the optimal state during motion tracking. 
For example, due to error accumulation, human global motion cannot be faithfully estimated
in the IMU-only branch.
%
On the other hand, the global motion information is directly measured by the visual signals in the other branch.
%
By our hidden state feedback technique, this error accumulation problem is solved as the hidden state in the IMU-only branch is calibrated by the other branch.
Specifically, instead of frequently updating the states or assuming a fixed time interval, our algorithm dynamically detects the need for hidden state updates and performs the strategy only when required.
This guarantees the high efficiency of our algorithm.
Furthermore, instead of calibrating one branch using the other, we use the final fused results, which is better than either of the two branches.
This effectively forms a feedback loop where the final combined result is used for hidden state updates in the two coordinate systems.
In summary, the hidden state feedback mechanism effectively and efficiently achieves information interchange between dual coordinate systems.

In this paper, our main contributions are:
\begin{itemize}
    \item An accurate and robust approach that fuses monocular images with sparse IMU signals for real-time human motion capture.
    \item A dual coordinate strategy that makes the neural network better learn from the inertial signals in different cases.
    \item A hidden state feedback mechanism that leverages the combined results in the loop by using them to improve the performance of individual components.
\end{itemize}
%
%
\section{Related Work}
%

\subsection{Vision-based Mocap Methods}
In vision-based mocap, the approaches using multi-view input have achieved remarkable results \cite{harvesting17, CrossVF, LearnableTO, remelli2020lightweight, chun2023learnable, reddy2021tessetrack}, but the system is heavy, and the recording space is limited. 
Some works use a single monocular camera, and they represent 3D human poses by skeletons \cite{pavllo20193d,sharma2019monocular,xu2020deep,zhen2020smap}, i.e., the 3D positions of all the body joints, which cannot represent the body shape and the exact 3D motion between two recorded image frames.
So some other works use human body parametric models, such as SMPL \cite{SMPL}, and thus the shape and pose parameters of the models are used to represent the body shape and the rotations of the body joints.
%
%
To obtain the parameters, the optimization-based methods \cite{Lassner, Bogo:ECCV:2016,xiang2019monocular,pavlakos2018learning} fit the parametric body models to 2D observations, while the regression-based methods \cite{kolotouros2019convolutional, hmr, kanazawa2019learning, VIBE, zanfir2021neural, minimal, zhang2022mixste, PhysAware, li2022cliff} train neural networks to estimate the model parameters directly from the input or some low-level features. 
%
%
%
In general, vision-based methods are severely affected by the invisibility issue caused by challenging lighting or occlusions.
Recent works, such as ROMP\cite{ROMP}, PARE\cite{PARE} 
have been devoted to better handling this problem, using a collision-aware representation or part-guided attention mechanism.
Other works \cite{yuan2022glamr,BEV,huang2022occluded} also focus on solving occlusions, and they design unique mechanisms, such as a deep generative motion infiller or a hypothetical bird's eye view.
%
%
Although these methods have handled occlusions to a certain extent, they still have difficulties handling extreme occlusions or cases of people out of the camera view.
\subsection{IMU-based Mocap Methods}
Unlike vision-based methods, inertial sensor-based methods are unaffected by challenging lighting, occlusions, and camera view limitations.  
Commercial inertial mocap solutions such as Xsens \cite{schepers2018xsens} have achieved high accuracy but require dense sensors (17 IMUs), making them inconvenient and expensive.  
Recently, mocap with sparse sensors has drawn much more attention.
%
%
%
%
%
SIP \cite{SIP} successfully reduces the number of sensors to 6, but is limited by the speed of the optimization-based approach and thus could not support real-time motion capture.
\cite{DIP, RNN-Ensemble} use recurrent neural networks to achieve real-time pose estimation with sparse IMUs for the first time, but they cannot estimate the global translation of people.  
TransPose \cite{TransPose} combines supporting-foot deduction and network prediction, and achieves real-time pose and global translation estimation with only 6 IMUs.  
\cite{schreiner2021global} focuses on the real-time estimation of precise position based on pose and orientation data.
Later, PIP \cite{PIP} introduces physical constraints \cite{physCap} to achieve higher accuracy, and a novel RNN initialization method to resolve the action ambiguity of long time sitting or standing still.
Meanwhile, TIP \cite{TIP} proposes stationary body points to mitigate the effects of drift.  
Additionally, \cite{jiang2022avatarposer,winkler2022questsim,ye2022neural3points,aliakbarian2022flag} use VR devices on the user's head and hands to estimate full-body poses.
However, due to the pose ambiguity of the sparse IMU setting and the sensor error accumulation, these methods suffer from jittery or drift in motion estimation.
%
\subsection{Vision-inertia Fusion Methods}
Depending on the type of visual input, the vision-inertia fusion methods can be divided into three categories: fusing multi-view video with IMUs \cite{pons2010multisensor,vonmarcardponsmollPAMI16,TotalCapture,malleson2020real,RealTimeFM,gilbert2019fusing,FusingWI,moniruzzaman2021wearable,bao2022fusepose,huang2020deepfuse}, fusing monocular RGBD video with IMUs \cite{helten2013real,zheng2018hybridfusion}, and fusing monocular RGB video with IMUs \cite{VIP,HybridCap,kaichi2020resolving,henschel2020accurate,henschel2019simultaneous}.
%
%
In the former two categories, some methods use optimization techniques by defining and minimizing energy functions that are correlated with both visual and IMU features \cite{vonmarcardponsmollPAMI16, RealTimeFM,pons2010multisensor,malleson2020real}.  
Other methods use dual-stream networks to estimate the pose from IMU and vision separately and then combine them \cite{TotalCapture,gilbert2019fusing}.  
However, the combination sometimes is not sufficient and may involve the errors of the two results together.
To solve this problem, \cite{FusingWI} proposes to fuse information at an earlier stage.  
%
%
FusePose \cite{bao2022fusepose} combines the IMU and visual information in the training stage so that they can adaptively compensate for each other.
%
%
%
For the methods combining monocular RGB with IMUs, the task is more challenging as monocular RGB contains less information than RGBD or multi-view RGB.
\cite{henschel2020accurate,henschel2019simultaneous} only estimate the global root translation.
\hl{Both \cite{kaichi2020resolving} and \cite{VIP} are involved in estimating body poses, but while \cite{kaichi2020resolving} necessitates dense IMUs, \cite{VIP} functions as an offline technique.}
Recently, one concurrent work HybridCap \cite{HybridCap} combines 4 IMUs and the 2D keypoint positions in the image domain to estimate the pose and translation.
They consider the situation that the two kinds of inputs are both available and they fail when the visual information is unavailable, as it is difficult for 4 IMUs to perform mocap.
We consider the visual and IMU fusion task in a different manner that we do not want the fused system to suffer the inherent limitation of visual motion capture, i.e., we handle the situation that the visual input is bad (challenging lighting, severe occlusion, or human out of viewing field).
%
By this design, we achieve the first real-time system that combines monocular RGB and sparse IMUs to achieve robust and accurate motion capture even when the visual signal is unavailable or severely polluted.
%
\begin{figure}
  \includegraphics[width=\linewidth]{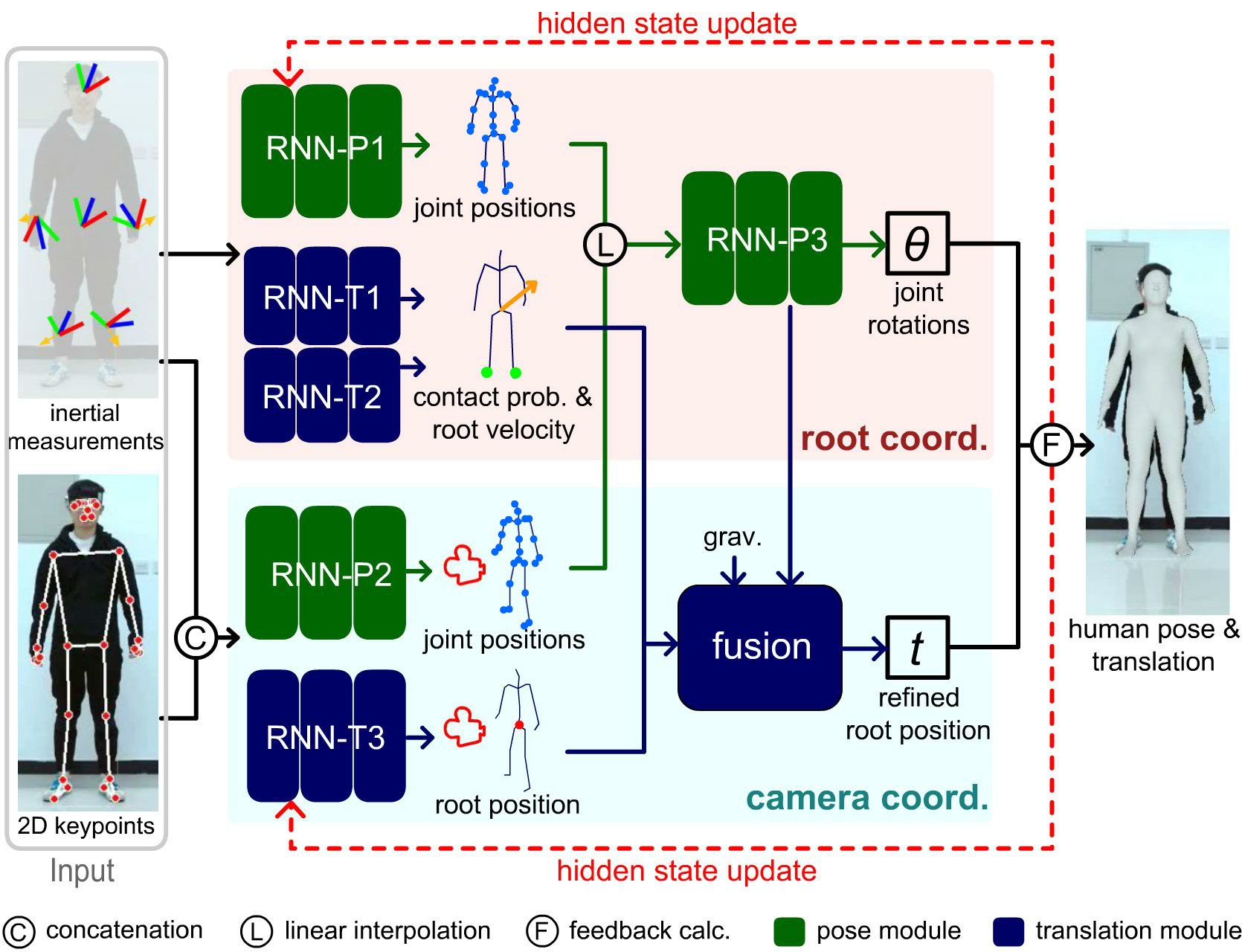}
  \caption{Overview of our method. The inputs are inertial measurements of 6 IMUs and image domain 2D keypoints obtained by an off-the-shelf 2D human pose estimator. The outputs are real-time human pose and translation. We adopt a dual coordinate strategy, where we estimate joint positions and global movements in both the human's root and camera coordinate systems and we fuse the results from two coordinate systems to get the human pose and translation. To enable the information interchange between the two coordinate systems and make them mutually help each other, we leverage a hidden state feedback mechanism to update the hidden state of the two branches using the final fused results. 
  }
  \label{fig:pipeline}
\end{figure}
\section{Method}
%
As shown in Fig. \ref{fig:pipeline}, the inputs of our method consist of sequential 2D joint detections of a moving subject from monocular images, as well as the accelerations and orientations of 6 IMUs mounted on the left and right forearms, left and right lower legs, head, and pelvis of the subject.
The output of our system is the 3D motion of the subject, represented on the kinematic model of SMPL \cite{SMPL}.
Note that a calibration step is performed to calibrate the relative orientation between the IMUs' coordinate system and the camera coordinate system, detailed in the supplementary materials.
Given this calibration, we can align the two input modalities by transforming either one to the coordinate system of the other.
In the following, we introduce our dual coordinate strategy which estimates both the global motion and local pose of the performer (Sec.~\ref{subsubsec:dual-coordinate}) as well as the hidden state feedback to make the estimation more robust and accurate (Sec.~\ref{sec:feedback}).
\subsection{Dual Coordinate Strategy}\label{subsubsec:dual-coordinate}
Deep learning has shown its great power in vision and graphic tasks.
How to represent its input and output significantly influence its performance in real applications.
In our deep learning-based technique, we propose to use two coordinate systems to represent our input in different situations and fuse them together to pursue a better estimation.
Our key observation is that the visual signals contain both the global and local motion of the performer in the camera coordinate system while the inertial signals majorly contain the local body movement information but the global motion information always drifts due to the inertial error accumulation.
So when the two modalities are both available, the camera coordinate system is more adequate to represent the input.
On the other hand, when the visual signals are not available (due to occlusions, extreme lighting, or out-of-camera view) and only the inertial signals are available, the root coordinate system is adopted because the same poses may differ significantly from different camera views but are precisely the same from a local view.
Moreover, the IMU-estimated root velocity in the root coordinate system is important to estimate the absolute position of the tracking subject when visual clues are unavailable.
%
%
%
%
%
As we have disentangled the local pose and global translation estimation, we introduce our dual coordinate strategy individually for the two tasks.
%
%
\subsubsection{Local Pose Estimation}\label{subsubsec:Pose}
We first describe local pose estimation.
To thoroughly learn the human motion prior, our algorithm first estimates joint 3D positions and then estimates joint rotations (i.e., local pose).
The joint position estimation task is solved under dual coordinates adaptive for different use cases (demonstrated by RNN-P1 for the root coordinate and RNN-P2 for the camera coordinate in Fig.~\ref{fig:pipeline}).
Then, we use linear interpolation to fuse the two results from the two coordinates and estimate the joint rotations under the root frame by inverse kinematics (demonstrated by RNN-P3).
%
\par
Specifically, when estimating joint positions in the root coordinate system, we transform the inertial measurements into the root coordinate by the IMU mounted on the pelvis, obtaining a concatenated input vector as the input of RNN-P1.
The input vector is denoted as $\boldsymbol{x}_{\mathrm{r}}=\left[\boldsymbol{a}_{\mathrm{larm}},\cdot\cdot\cdot,\boldsymbol{a}_{\mathrm{root}},\boldsymbol{R}_{\mathrm{larm}},\cdot\cdot\cdot,\boldsymbol{R}_{\mathrm{root}}\right]\in\mathbb{R}^{(3+9)n}$ where $\boldsymbol{a}\in\mathbb{R}^3$ represents the acceleration, $\boldsymbol{R}\in\mathbb{R}^{3\times3}$ represents the rotation, the subscript $\mathrm{r}$ denotes the root coordinate system, and $n=6$ represents the number of IMUs.
RNN-P1 outputs the root-relative coordinates of all joints as $\boldsymbol{p}_{\mathrm{r}}^{\mathrm{e}}\in\mathbb{R}^{3J}$, where $J$ represents the number of joints, and the superscript $\mathrm{e}$ indicates the estimation.
The loss function used to train  RNN-P1 is defined as:
\begin{equation}
    \mathcal{L}_{P1}=\Vert \boldsymbol{p}_{\mathrm{r}}^{\mathrm{e}}-\boldsymbol{p}_\mathrm{r}^{\mathrm{GT}} \Vert_2^{2},
\end{equation}
where superscript $^{\mathrm{GT}}$ indicates the ground truth.
\par
When estimating joint positions in the camera coordinate system, we first transform the inertial measurements into the camera coordinate system.
We then reproject the 2D keypoints, detected by MediaPipe \cite{mediapipe}, onto the $Z$ = 1 plane using camera intrinsics to generalize to various camera settings.
After this, we get $\boldsymbol{p}_{\mathrm{2d}}\in\mathbb{R}^{2J^{'}}$ where $J'$ represents the number of detected keypoints from MediaPipe.
Next, we perform root normalization to represent the keypoints as root-relative 2D key points and the absolute position of the root in the camera coordinate system.
After the root normalization, we obtain $\boldsymbol{p}_{\mathrm{2d}}^{'}$.
The input of RNN-P2 can be represented as $[\boldsymbol{x}_{\mathrm{c}},\boldsymbol{p}_{\mathrm{2d}}^{'},\boldsymbol{\sigma}]\in\mathbb{R}^{(3+9)n+3J^{'}}$, where $\boldsymbol{x}_{\mathrm{c}}\in\mathbb{R}^{(3+9)n}$ denotes the IMU measurements in the camera coordinate system, and $\boldsymbol{\sigma}\in\mathbb{R}^{J^{'}}$ represents the keypoints' corresponding confidence scores given by MediaPipe.
The output of RNN-P2 is root-relative joint positions in the camera coordinate system, denoted as $\boldsymbol{p}_{\mathrm{c}}^{\mathrm{e}}\in\mathbb{R}^{3J}$.
The loss function for RNN-P2 is the same as the loss function for RNN-P1.
\par
To fuse the results in the two coordinate system, we first transform $\boldsymbol{p}_{\mathrm{c}}^{\mathrm{e}}\in\mathbb{R}^{3J}$ into the root coordinate system.
Then, we can fuse them to get the final root-relative coordinates of all joints denoted as $\boldsymbol{p}_{\mathrm{r}}$.
We use the average visual confidence of all keypoints $\sigma_{\mathrm{mean}}$ to determine the fusing process.
Specifically, if the average confidence $\sigma_{\mathrm{mean}}$ is sufficiently high (i.e., visual information is of good quality), we take the results from the camera coordinate frame;
if $\sigma_{\mathrm{mean}}$ is significantly low (i.e., visual information is unreliable or absent), we take the joints estimated from solely IMUs.
Otherwise, a linear interpolation is performed between the two results to smooth the process.
We experimentally set the lower bound and upper bound to 0.7 and 0.8, respectively.
Finally, RNN-P3 takes the concatenated vector $[\boldsymbol{p}_{\mathrm{r}},\boldsymbol{x}_{\mathrm{r}}]$ as input to estimate the joint rotations in 6D representation \cite{zhou2019continuity}, denoted as $\boldsymbol{\varphi}\in\mathbb{R}^{6J}$.
We can convert $\boldsymbol{\varphi}$ to axis-angle representation and obtain $\boldsymbol{\theta}_\mathrm{r}$.
The loss of RNN-P3 is defined as:
\begin{equation}
\mathcal{L}_{P3}=\lambda_{\mathrm{rot}}\mathcal{L}_{\mathrm{rot}}+\lambda_{\mathrm{pos}}\mathcal{L}_{\mathrm{pos}}.
\end{equation}
$\mathcal{L}_{\mathrm{rot}}$, which constrains the estimated joint rotations, is defined as:
\begin{equation}  
\mathcal{L}_{\mathrm{rot}}=\Vert \boldsymbol{\varphi}-\boldsymbol{\varphi}^{\mathrm{GT}} \Vert_2^{2}.
\end{equation}
$\mathcal{L}_{pos}$, which constrains the joint positions after forward kinematics, is defined as: 
\begin{equation}  
\mathcal{L}_{pos}=\Vert \mathrm{FK}\left(\boldsymbol{\theta}_{\mathrm{r}}\right)-\mathrm{FK}\left(\boldsymbol{\theta}_\mathrm{r}^{\mathrm{GT}}\right) \Vert_2^{2},
\end{equation}
where $\mathrm{FK}(\cdot)$ is the forward kinematics function that calculates the position of the joints.
We experimentally set the parameters $\lambda_{\mathrm{rot}}=1$ and $\lambda_{\mathrm{pos}}=100$.
\hl{By multiplying the orientation measurement of the root IMU with the calibrated camera extrinsic, joint rotations $\boldsymbol{\theta}\mathrm{r}$ can be transformed into the camera coordinate system, resulting in $\boldsymbol{\theta}\mathrm{c}$.}
\begin{table*}[t]
\caption{Quantitative comparisons with state-of-the-art methods ROMP, PARE, TIP, PIP, HybridCap, and VIP. We show results on 3DPW (in the wild pose), 3DPW-OCC (in the wild occluded pose), AIST++ (challenging motion), and TotalCapture (with subject out-of-view scenarios) datasets.}\label{tab:allcmp}
\resizebox{\textwidth}{!}{
\begin{tabular}{c|ccc|ccc|cccc|cccc}
\hline
\multicolumn{1}{c|}{\multirow{2}{*}{Method}} & \multicolumn{3}{c|}{3DPW}                                                                  & \multicolumn{3}{c|}{3DPW-OCC}  
                                             & \multicolumn{4}{c|}{AIST++}                                                                 & \multicolumn{4}{c}{TotalCapture}                                    \\ 
                                             \cline{2-15} 
\multicolumn{1}{l|}{}                        & \multicolumn{1}{c}{MPJPE} & \multicolumn{1}{c}{PA-MPJPE} & PVE              & \multicolumn{1}{c}{MPJPE} & \multicolumn{1}{c}{PA-MPJPE} & PVE
                                             & \multicolumn{1}{c}{MPJPE} & \multicolumn{1}{c}{PA-MPJPE} & \multicolumn{1}{c}{PVE} &TE & \multicolumn{1}{c}{MPJPE} & \multicolumn{1}{c}{PA-MPJPE} & \multicolumn{1}{c}{PVE} &TE \\ 
\hline
ROMP                                         & \multicolumn{1}{c}{91.3}       & \multicolumn{1}{c}{54.9}    & \multicolumn{1}{c|}{108.3}  & \multicolumn{1}{c}{-}          & \multicolumn{1}{c}{-}             & \multicolumn{1}{c|}{-}   
                                             & \multicolumn{1}{c}{90.3}       & \multicolumn{1}{c}{60.0}          & \multicolumn{1}{c}{128.1}    &16.91                    & \multicolumn{1}{c}{145.4}      & \multicolumn{1}{c}{64.6}          & \multicolumn{1}{c}{184.5}   &60.31 \\ 
PARE                                         & \multicolumn{1}{c}{82.0}       & \multicolumn{1}{c}{50.9}    & \multicolumn{1}{c|}{97.9}   & \multicolumn{1}{c}{90.5}       & \multicolumn{1}{c}{56.6}          & \multicolumn{1}{c|}{107.9}   
                                             & \multicolumn{1}{c}{83.7}      & \multicolumn{1}{c}{50.9}     & \multicolumn{1}{c}{116.5}   & -                   & \multicolumn{1}{c}{143.5}     & \multicolumn{1}{c}{60.9}           & \multicolumn{1}{c}{196.8}   & -       \\ 
TIP                                          & \multicolumn{1}{c}{82.5}       & \multicolumn{1}{c}{58.2}    & \multicolumn{1}{c|}{109.9}  & \multicolumn{1}{c}{100.5}      & \multicolumn{1}{c}{68.7}          & \multicolumn{1}{c|}{131.6}  
                                             & \multicolumn{1}{c}{85.1}      & \multicolumn{1}{c}{62.1}     & \multicolumn{1}{c}{115.5}         &69.52                     & \multicolumn{1}{c}{69.3}      & \multicolumn{1}{c}{35.7}            & \multicolumn{1}{c}{88.9}   &56.18   \\ 
PIP                                          & \multicolumn{1}{c}{78.0}       & \multicolumn{1}{c}{49.8}    & \multicolumn{1}{c|}{100.0}  & \multicolumn{1}{c}{97.8}       & \multicolumn{1}{c}{66.0}          & \multicolumn{1}{c|}{126.1}    
                                             & \multicolumn{1}{c}{87.1}      & \multicolumn{1}{c}{62.0}     & \multicolumn{1}{c}{116.5}         &45.17                     & \multicolumn{1}{c}{49.1}      & \multicolumn{1}{c}{34.6}            & \multicolumn{1}{c}{66.0}   &43.77   \\
Hybridcap                                    & \multicolumn{1}{c}{72.1}       & \multicolumn{1}{c}{-}       & \multicolumn{1}{c|}{-}      & \multicolumn{1}{c}{-}          & \multicolumn{1}{c}{-}             & \multicolumn{1}{c|}{-}    
                                             & \multicolumn{1}{c}{33.3}  & \multicolumn{1}{c}{-}         & \multicolumn{1}{c}{-}    & -                   & \multicolumn{1}{c}{-}         & \multicolumn{1}{c}{-}               & \multicolumn{1}{c}{-}      & -\\
VIP                                          & \multicolumn{1}{c}{-}       & \multicolumn{1}{c}{-}       & \multicolumn{1}{c|}{-}      & \multicolumn{1}{c}{-}          & \multicolumn{1}{c}{-}             & \multicolumn{1}{c|}{-}    
                                             & \multicolumn{1}{c}{-}  & \multicolumn{1}{c}{-}         & \multicolumn{1}{c}{-}    & -                   & \multicolumn{1}{c}{-}         & \multicolumn{1}{c}{39.6}               & \multicolumn{1}{c}{-}      & -\\ 
Ours                             & \multicolumn{1}{c}{\textbf{55.0}}& \multicolumn{1}{c}{\textbf{38.9}}& \multicolumn{1}{c|}{\textbf{71.8}} & \multicolumn{1}{c}{\textbf{77.9}}& \multicolumn{1}{c}{\textbf{53.1}}& \multicolumn{1}{c|}{\textbf{97.5}}         
                    & \multicolumn{1}{c}{\textbf{33.1}}& \multicolumn{1}{c}{\textbf{24.0}}& \multicolumn{1}{c}{\textbf{43.2}}   & \textbf{9.92}       & \multicolumn{1}{c}{\textbf{48.7}}& \multicolumn{1}{c}{\textbf{33.5}} & \multicolumn{1}{c}{\textbf{63.4}}     & \textbf{23.52} \\
\hline
\end{tabular}}
\end{table*}
\subsubsection{Global Translation Estimation}\label{subsubsec:Tran}
The translation estimation differs from pose estimation because pure inertial-based methods suffer from large drifts due to accumulated errors, as demonstrated in~\cite{EgoLocate}.
Thus, we adopt a slightly different method for translation estimation.
When the visual information is not reliable, we follow TransPose~\cite{TransPose} to use IMU only to estimate the translation. 
%
%
%
On the other hand, when visual information is reliable, we directly estimate the absolute position of the human in the camera coordinate system (as shown by RNN-T3).
Then we use the complementary filter algorithm to fuse the two results depending on the reliability of the visual signals.
%
%
%
%
%
%
\par
To be more specific, RNN-T1 takes inertial measurements $\boldsymbol{x}_{\mathrm{r}}$ as input to estimate foot-ground contact probability $s$.
On the other hand, RNN-T2 takes $\boldsymbol{x}_{\mathrm{r}}$ as input and estimates the root velocity $\boldsymbol{{v}_{\mathrm{r}}^{\mathrm{e}}}\in\mathbb{R}^{3}$ in the root coordinate system.
To train the models (RNN-T1 and RNN-T2), we utilize the same loss function as TransPose~\cite{TransPose}.
We can calculate the foot velocity $\boldsymbol{v}_{\mathrm{r}}^\mathrm{f}\in\mathbb{R}^{3}$ using the estimated pose $\boldsymbol{\theta}_{\mathrm{r}}$ as:
\begin{equation}
\boldsymbol{v}_{\mathrm{r}}^\mathrm{f}=\frac{1}{\Delta t}(\mathrm{GF}(\mathrm{FK}(\boldsymbol{\theta}_{\mathrm{r}}^{(k)}-\mathrm{GF}(\mathrm{FK}(\boldsymbol{\theta}_{\mathrm{r}}^{(k-1)})),
\end{equation}
where $\mathrm{GF}(\cdot)$ represents the function that retrieves the supporting foot position from all joint positions, $\boldsymbol{\theta}_{\mathrm{r}}^{(k)}$ represents the joint rotations in the root coordinate system at frame $k$, and $\Delta t$ is the time interval between frames.
%
%
If the foot-ground contact probability $s$ is smaller than a threshold $0.7$, we take the estimated root velocity $\boldsymbol{{v}_{\mathrm{r}}^{\mathrm{e}}}$.
Otherwise, we derive the root velocity from the foot velocity $\boldsymbol{v}_{\mathrm{r}}^\mathrm{f}$ following TransPose.
We denote the final root velocity as  $\boldsymbol{{v}_{\mathrm{r}}}$ in the root coordinate system and transform it into the camera coordinate system as $\boldsymbol{{v}_{\mathrm{c}}}$.
Notice that if we accumulate $\boldsymbol{{v}_{\mathrm{c}}}$ to estimate the root position, the result is prone to low-frequency drift due to the error accumulation.
\par
For RNN-T3, we concatenate the inertial measurements in the camera coordinate $\boldsymbol{x}_{\mathrm{c}}$, the canonicalized keypoints $\boldsymbol{p}_{\mathrm{2d}}$, and the corresponding confidence $\boldsymbol{\sigma}$ to estimate the global root position $\boldsymbol{t}_{\mathrm{c}}^{\mathrm{e}}$ of the subject in the camera coordinate system. 
The loss function used to train this model is denoted as:
\begin{equation}
\mathcal{L}_{T3}=\Vert \boldsymbol{t}_{\mathrm{c}}^{\mathrm{e}}-\boldsymbol{t}_\mathrm{c}^{\mathrm{GT}} \Vert_2^{2}.
\end{equation}
Also note that due to the noise in the off-the-shelf 2D keypoint detector and depth ambiguity, this position estimation somehow contains high-frequency noise.
\par
%
Next, we fuse the estimated root position $\boldsymbol{t}_{\mathrm{c}}^{\mathrm{e}}$ and the global velocity $\boldsymbol{{v}_{\mathrm{c}}}$ to obtain the final position $\boldsymbol{{t}_{\mathrm{c}}}$ in the camera coordinate system.
%
For the fusion process, we utilize the average keypoint confidence $\sigma_{\mathrm{mean}}$ as a reference to fuse the results using the complementary filter algorithm.
%
%
%
The fusion algorithm can be mathematically defined as:
\begin{equation}
    \boldsymbol{t}_{\mathrm{c}}^{(k)} = (1-\alpha_k)\boldsymbol{t}_{\mathrm{c}}^{(k-1)} + (1-\alpha_k)\boldsymbol{v}_{\mathrm{c}}^{(k)}\Delta t + \alpha_k\boldsymbol{t}_{\mathrm{c}}^{\mathrm{e}(k)},
\end{equation}
where $k$ represents the $k$th frame, $\Delta t$ is the time interval between frames, and $\alpha_k=0.05\sigma_{\mathrm{mean},k}$ is the coefficient dynamically determined by the average keypoint confidence at the $k$th frame $\sigma_{\mathrm{mean},k}$.
Intuitively, when visual confidence is low, we directly add the relative translation estimated solely by IMUs to the previous root position.
When visual confidence is high, after updating the root position with the IMU-based relative translation, we perform an additional correction step leveraging the estimated root position in the camera frame.
\par
Finally, we optimize the joint rotations $\boldsymbol{\theta}_{\mathrm{c}}$ and the global position $\boldsymbol{t}_{\mathrm{c}}$ to minimize the reprojection error and get the final motion output.
%
%
Our energy function is defined as:
\begin{equation}
\boldsymbol{E}(\boldsymbol{\theta}_{\mathrm{c}},\boldsymbol{t}_{\mathrm{c}})=\lambda_{\mathrm{2D}}\boldsymbol{E}_{\mathrm{2D}}+\lambda_{\mathrm{3D}}\boldsymbol{E}_{\mathrm{3D}}+\lambda_{\mathrm{prior}}\boldsymbol{E}_{\mathrm{prior}}+\lambda_{\mathrm{angle}}\boldsymbol{E}_{\mathrm{angle}}+\lambda_{\mathrm{ori}}\boldsymbol{E}_{\mathrm{ori}}.
\end{equation}
$\boldsymbol{E}_{\mathrm{2D}}$ enforces the final joint positions reprojects onto 2D close to the 2D keypoints detected by MediaPipe. 
$\boldsymbol{E}_{\mathrm{3D}}$ enforces the final joint positions close to the predicted one.
Following \cite{smplify}, $\boldsymbol{E}_{\mathrm{prior}}$ penalizes the unrealistic human pose using a Geman-McClure \cite{geman-mcclure} error function.
Following \cite{smplify}, $\boldsymbol{E}_{\mathrm{angle}}$ penalizes unnatural bending of the knees and elbows.
And $\boldsymbol{E}_{\mathrm{ori}}$ enforces the rotations of joints with IMU mounted close to the inertial measurements.
We experimentally set $\lambda_{\mathrm{2D}}=1$, $\lambda_{\mathrm{3D}}=1$, $\lambda_{\mathrm{prior}}=0.1$, $\lambda_{\mathrm{prior}}=15.2$, and $\lambda_{\mathrm{ori}}=0.5$.
We use the L-BFGS \cite{lbfgs} algorithm to solve this optimization with 1 iteration.
\subsection{Hidden State Feedback Mechanism}\label{sec:feedback}
The aforementioned dual coordinate strategy can preserve the final result by utilizing the benefits of both branches.
However, the two branches do not mutually benefit from each other.
For RNN-P1, inspired by \cite{PIP},  utilizing a constant initial hidden state for the RNN training is incorrect.
This is because the subject may initiate from different poses, and if the initial state is incorrect, the network can not learn how to change its hidden state according to the changing input signals.
%
%
In the case of RNN-T3, when the subject is occluded for a while and reappears in the camera view, the network cannot immediately generate accurate root positions as its hidden states have not been correctly updated during the occlusion.
For RNN-P2, RNN-T1, and RNN-T2, we consider them to be less temporally dependent, i.e., these models primarily rely on the input of the current frame rather than the historical information.
This consideration motivates us to develop a hidden state feedback mechanism that can enhance the two branches.
\par
Considering the difference between RNN-P1 and RNN-T3, we propose different feedback schemes for each.
For RNN-P1, as mentioned earlier, a constant initialization of the hidden state in the RNN leads to inaccuracies.
PIP utilizes the ground-truth pose to initialize the hidden state of the RNN.
This is possible for the first frame but not for the following frames, so after tracking a lot of frames, the initialization has no contribution to the tracking anymore.
%
%
On the other hand, we have the visual input, and thus we could use the joint positions $\boldsymbol{p}_{\mathrm{c}}^{\mathrm{e}}$ obtained from RNN-P2 to initialize RNN-P1 in the middle of the tracking in case the visual signals are good enough to estimate an accurate pose.
%
%
By leveraging the visual information, this mechanism provides a more flexible and adaptive initialization scheme for RNN-P1 during the tracking.
This mechanism significantly better distinguishes ambiguous motions, such as standing still and sitting still, because they share identical inertial measurements and the correct historical information is the key to solving the ambiguity.
\par
For RNN-T3, since its visual input may be meaningless (e.g. when the performer is out of the camera view), we use the final fused result to update its hidden state if the mean visual confidence of all joints $\sigma_\mathrm{mean}$ is less than the predefined lower bound.
To be specific, we use synthesized 2D keypoints $\boldsymbol{p}_\mathrm{2d}$ instead of the unreliable MediaPipe-detected keypoints under these conditions.
We synthesize $\boldsymbol{p}_\mathrm{2d}$ as:
\begin{equation}
    \boldsymbol{p}_\mathrm{2d} = \mathrm{\Pi}\left(\mathrm{FK}\left(\boldsymbol{\theta}_\mathrm{c}\right)+\boldsymbol{t}_\mathrm{c}\right),
\end{equation}
where $\mathrm{\Pi}$(·) represents the projection onto $Z=1$ plane from 3D space.
Then, we utilize $[\boldsymbol{x}_{\mathrm{c}},\boldsymbol{p}_{\mathrm{2d}},\boldsymbol{\sigma}]$ to run RNN-T3 to update its hidden state.
%
Due to this hidden state feedback, when $\sigma_\mathrm{mean}$ increases, RNN-T3 will recover immediately to give a result that is both reasonable and consistent with the previous output, avoiding sudden changes.
The effectiveness of this feedback mechanism is noticeable when the subject comes back after going outside of the camera view, especially when the re-entry position is different from the going-out position. We will show this in the following section.
\begin{figure}
\includegraphics[width=\linewidth]{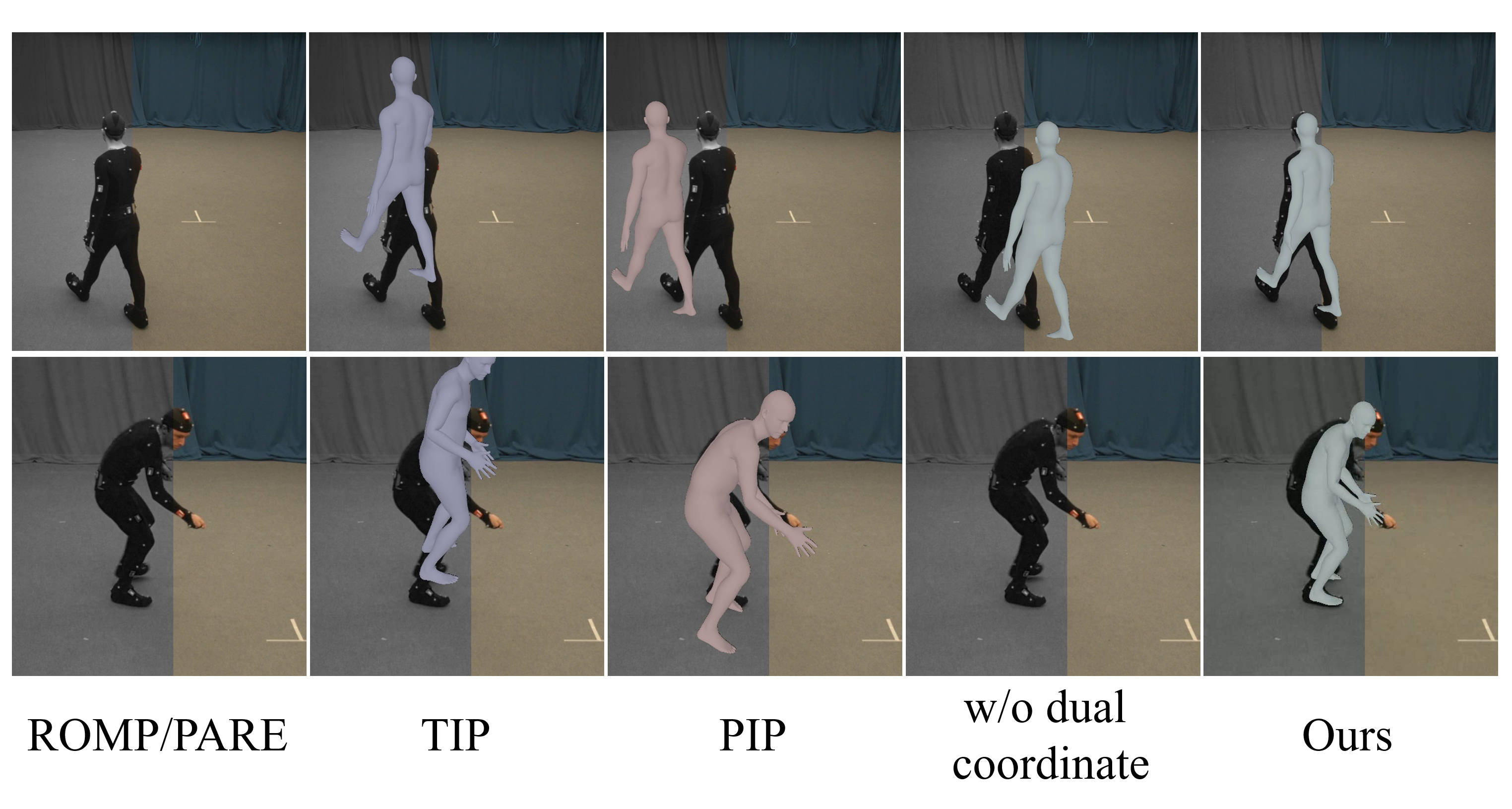}
  \centering
  \caption{
   Qualitative results on the TotalCapture dataset that the subjects are out of the camera view. Visual signals are not available in the gray regions. From left to right: ROMP/PARE results, TIP results, PIP results, our method without the dual coordinate strategy, and our results.}
  \label{fig:comparison2}
\end{figure}
\section{Experiments}
In this section, we first provide more details of our implementation and the datasets (Sec.~\ref{sec:implementation}).
Then we compare our method with previous works (Sec.~\ref{sec:comparisons}).
Next, we evaluate our key components (Sec.~\ref{sec:evaluations}).
Finally, we show more results and live demos in our accompanying video (Sec.~\ref{sec:moreResults}) and discuss our limitations (Sec.~\ref{sec:limitations}).
\subsection{Implementation Details}\label{sec:implementation}
All training and evaluation processes run on a computer with an Intel(R) Core(TM) i7-8700 CPU and an NVIDIA GTX2080Ti graphics card.
We train each recurrent neural network separately with its own loss (without using previous results from other RNNs) and connect them to form our entire method.
Following Transpose \cite{TransPose}, we add a gravity velocity to the translation output.
%

\paragraph{Datasets.}
We use the AIST++ dataset \cite{AIST++} and the AMASS dataset \cite{AMASS} for training.
Our model is trained on AIST++ using detected 2D keypoints and synthesized IMUs, and on AMASS using synthesized 2D keypoints and IMUs.
We perform our evaluations on TotalCapture \cite{TotalCapture}, AIST++ test split, 3DPW test split \cite{VIP}, and 3DPW-OCC \cite{VIP, ooh2020}.
Since the AIST++, 3DPW, and 3DPW-OCC datasets do not have IMU data, we utilize synthesized IMUs.
Please refer to our supplementary for the network architecture and more details about datasets.

\paragraph{Metrics.}
We use the following metrics to evaluate our method.
1) MPJPE is the mean per joint position error in mm.
2) PA-MPJPE is the Procrustes-aligned mean per joint position error in mm.
3) PVE is a per-vertex error in mm.
4) TE is a translation error that measures the absolute position error in cm.
Among these metrics, 1), 2), and 3) measure the pose accuracy, and 4) measures the global translation accuracy.
\subsection{Comparisons}\label{sec:comparisons}
\paragraph{Quantitative.}
We perform the comparisons with the state-of-the-art vision-based methods ROMP \cite{ROMP}, PARE \cite{PARE}, IMU-based methods TIP \cite{TIP}, PIP \cite{PIP} and combined method HybridCap \cite{HybridCap}, VIP \cite{VIP}.
PARE and VIP are both offline solutions.
Notice that PARE does not evaluate the global translation; it only estimates camera parameters that can project the subject onto the image.
For HybridCap and VIP, since they do not provide their codes, we compare them in the AIST++, 3DPW, and TotalCapture datasets with the numbers in the paper.
For all the previous methods, we use their default settings.
\par
\begin{figure*}
\includegraphics[width=\linewidth]{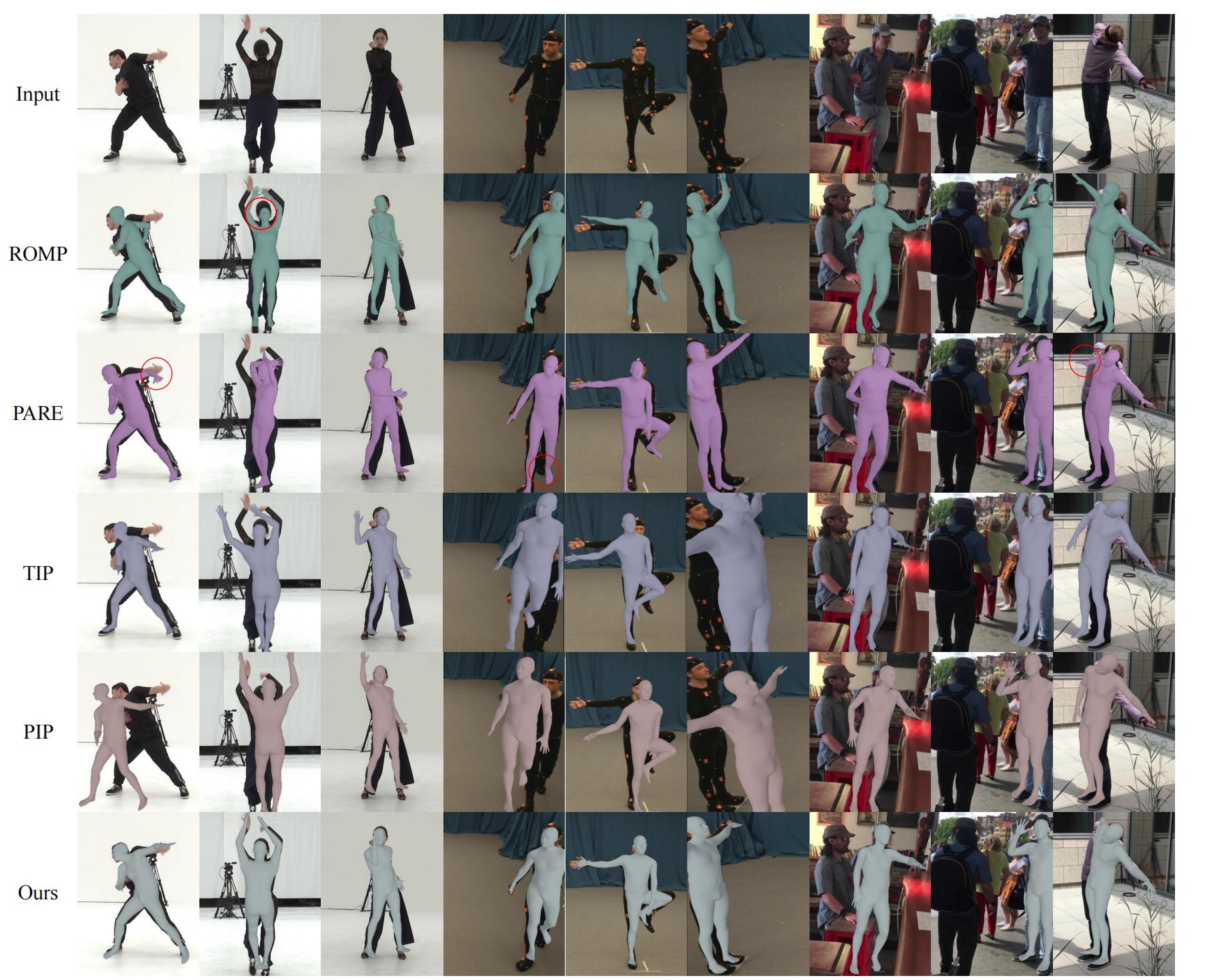}
  \centering
  \caption{
   Qualitative results on the AIST++ (columns 1-3), TotalCapture (columns 4-6), 3DPW, and the 3DPW-OCC Dataset (columns 7-9).}
  \label{fig:comparison}
\end{figure*}

The quantitative results are shown in Tab. ~\ref{tab:allcmp}.
Our method demonstrates significant improvements in pose and translation accuracy compared to previous works.
The TotalCapture dataset contains 8 camera views, and there are instances where the subject experiences severe occlusion or is out of the camera view.
Pure vision-based method ROMP and PARE fail in these cases.
With the help of our dual coordinate strategy and feedback mechanism, we achieve better results than ROMP and PARE.
We also reduce the pose and translation errors compared to pure IMU-based methods such as TIP, PIP, and the combined method VIP.
Besides, the AIST++ dataset comprises 9 camera views, and the subjects perform numerous challenging dancing motions, all captured within the camera view.
We significantly reduce pose and translation errors compared to ROMP, PARE, TIP, and PIP.
Despite HybridCap being designed explicitly for challenging poses, we consistently achieve superior results compared to it.
Finally, even without training on in-the-wild datasets, we still perform better on 3DPW and 3DPW-OCC, representing in-the-wild and occlusion scenarios.
\paragraph{Qualitative.}
In Fig. ~\ref{fig:comparison}, we present the qualitative results of our methods compared to ROMP, PARE, TIP, and PIP when the subject is within the camera view.
Vision-based method ROMP and PARE can better estimate the positions, but the reconstructed poses of end joints sometimes need to be corrected as these regions are small and thus have limited pixels to guide the vision-based methods.
On the other hand, IMU-based methods TIP and PIP are suffering drifts indicated by the slightly worse overlay.
%
%
Our approach leverages vision and IMU inputs more effectively than previous works through our dual coordinate strategy and hidden state feedback mechanism.
As a result, we can achieve better pose and translation estimation in challenging motion, occluded, and in-the-wild scenarios.
In Fig. ~\ref{fig:comparison2}, we show the qualitative comparison results when the subject is out of the camera view.
Each image's left part (denoted by the gray color) is regarded as out of the camera view in this experiment, and we show this part as a reference to evaluate different methods.
ROMP and PARE fail as the visual signal is not available.
PIP and TIP estimate plausible poses but suffer the drift problem.
Due to the hidden state feedback, our IMU mocap branch is corrected from time to time to reduce the drift, so it estimates accurate pose and much better translation when subjects are out of view.
%
%
\begin{table}[t]
\caption{Quantitative ablation study on the dual coordinate strategy, hidden state feedback mechanism, and optimization. We performed it on TotalCapture and 3DPW-OCC datasets as these two datasets contain cases where humans are occluded or out of the camera view, which are challenging and best demonstrate our key algorithms.}
\label{tab:ablitioncmp}
\resizebox{\columnwidth}{!}{
\begin{tabular}{c|cccc|ccc}
\hline
\multirow{2}{*}{Method} & \multicolumn{4}{c|}{TotalCapture}                       & \multicolumn{3}{c}{3DPW-OCC}                             \\ \cline{2-8} 
                        & \multicolumn{1}{c}{MPJPE} & \multicolumn{1}{c}{PA-MPJPE} & \multicolumn{1}{c}{PVE}      & TE      & \multicolumn{1}{c}{MPJPE} & \multicolumn{1}{c}{PA-MPJPE} & PVE \\ \hline
w/o dual coordinate     & \multicolumn{1}{c}{55.2}  & \multicolumn{1}{c}{37.9}     & \multicolumn{1}{c}{69.9}     & 37.09   & \multicolumn{1}{c}{81.1}  & \multicolumn{1}{c}{55.9}     & 102.0            \\ 
w/o feedback            & \multicolumn{1}{c}{49.5}  & \multicolumn{1}{c}{34.0}     & \multicolumn{1}{c}{64.2}     & 23.82   & \multicolumn{1}{c}{78.1}  & \multicolumn{1}{c}{53.4}     & 98.0   \\
w/o optimization   & \multicolumn{1}{c}{48.8} & \multicolumn{1}{c}{33.6} & \multicolumn{1}{c}{\textbf{63.1}} & 25.28 & \multicolumn{1}{c}{\textbf{75.7}}  & \multicolumn{1}{c}{\textbf{52.2}}     & \textbf{97.1}   \\
Ours            & \multicolumn{1}{c}{\textbf{48.7}}& \multicolumn{1}{c}{\textbf{33.5}}  & \multicolumn{1}{c}{{63.4}}& \textbf{23.52}   & \multicolumn{1}{c}{77.9}& \multicolumn{1}{c}{53.1}  & 97.5   \\ \hline
\end{tabular}}
\end{table}
\begin{figure}
  \includegraphics[width=\linewidth]{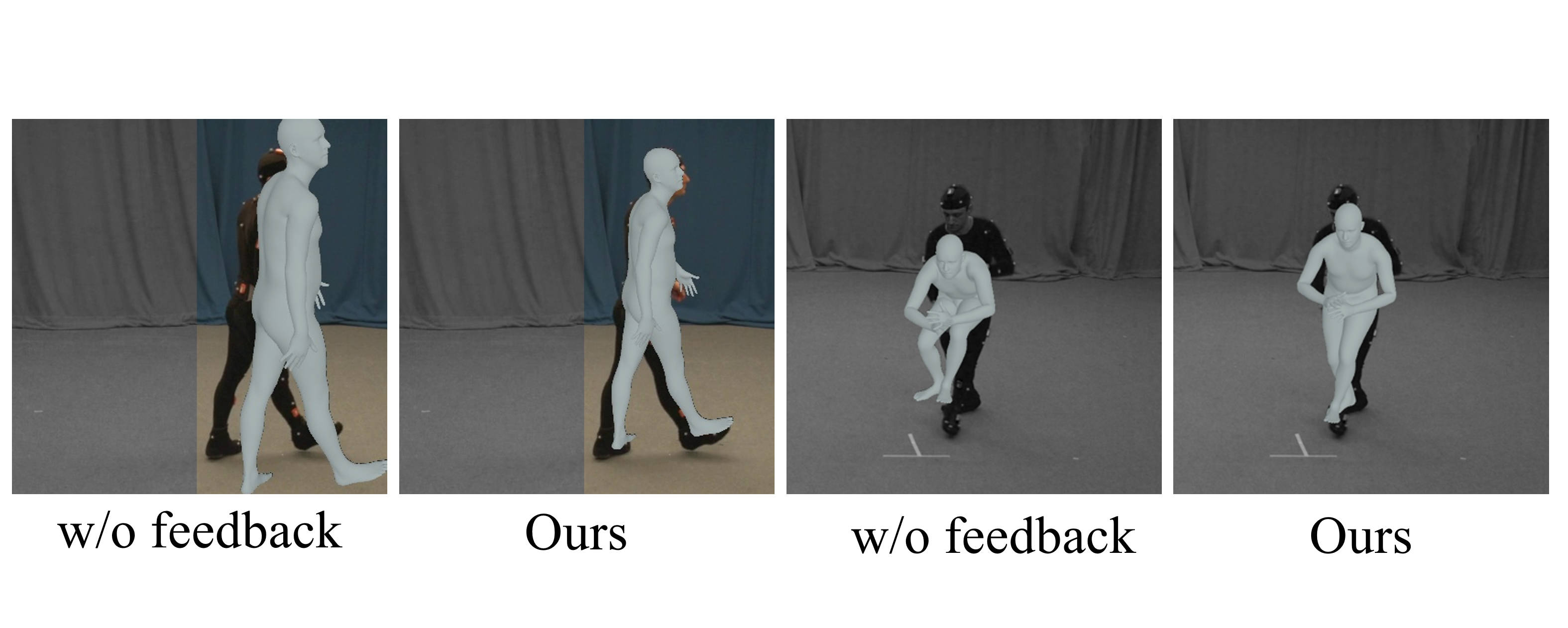}
  \centering
  \caption{
  Qualitative ablation study on the hidden state feedback mechanism. Visual signals are not available in the gray figure.}
  \label{fig:evaluation}
\end{figure}
\subsection{Evaluations}\label{sec:evaluations}
We evaluate three critical components of our method: the dual coordinate strategy, the hidden state feedback and the optimization.
\paragraph{Dual Coordinate Strategy.}
We compare our solution with a setting where all the estimations are in the camera coordinate.
Quantitative results are shown in Tab. ~\ref{tab:ablitioncmp}.
We can clearly see that the dual coordinate strategy makes significant performance improvement.
Qualitative results are also shown in Fig. ~\ref{fig:comparison2}.
Without our dual coordinate strategy, in the camera coordinate system, the method struggles to estimate the translation accurately, and thus it becomes more challenging to estimate the pose from the IMUs.  
\paragraph{Hidden State Feedback.}
To examine the effect of the feedback mechanism, we train two alternatives that do not update the hidden states of RNN-T3 and RNN-P1, respectively.
As discussed in the Method section, the feedback for RNN-T3 is most effective in determining translations when the subject is moving in again after moving out of the camera view.
Since TotalCapture and 3DPW-OCC contain this kind of motion, the translation error is increasing on TotalCapture and 3DPW-OCC in Tab. ~\ref{tab:ablitioncmp} for the former alternative.
To better demonstrate this, we pick a walking sequence in the TotalCapture dataset, and the result is shown in the left side of Fig. ~\ref{fig:evaluation}, where the method without feedback fails when the subject moves in the camera view at a different place from where the subject moved out.
The error is because the networks learn to avoid sudden translation changes.
So, without the feedback mechanism, the vision branch remembers where the subject moves out of the camera view. When the branch is activated again (the subject comes in again), it leans to estimate the translation near the remembered place.
However, with the feedback, the synthesized input $\boldsymbol{p}_{\mathrm{2d}}$ can update the hidden state of RNN-T3. On the condition that the IMU mocap works well when the subject is out of the camera view, RNN-T3 can be correctly restarted when the subject comes in again at a new place. 
\par
The feedback for RNN-P1 is most helpful in reconstructing the ambiguous poses whose IMU measurements are identical.
We additionally select a sequence from the TotalCapture dataset, and the corresponding results are presented on the right side of Fig. ~\ref{fig:evaluation}.
The gray-colored region in this figure is regarded as out of the camera view in this experiment (the subject is in the camera view at the beginning), and we show this region as a reference for better evaluation.
In the beginning, the subject is visible to the camera, so the combined branch updates the hidden states of the IMU branch, thus making it known that the subject is bending rather than sitting.
Then when the subject moved out of view, the method with the feedback still reconstructs a bending pose while the one without feedback solves a sitting pose as these two poses share similar IMU signals and sitting data is much more than the bending data in the training dataset.
%
%
\paragraph{Optimization.}
As shown in Tab. ~\ref{tab:ablitioncmp}, the final 1-iteration optimization primarily aims to improve the 2D overlay, and it has a slight effect on the pose result while reducing the translation error.
\hl{In the case of the 3DPW-OCC dataset, the pose error is larger when using optimization due to occlusion affecting the 2D detector.}
\subsection{More Results}\label{sec:moreResults}
We test our method on various motions, including moving out and re-entry the camera view, extreme lighting, severe occlusion (e.g., clothes, umbrellas, and boards), and challenging poses (e.g., rapid jumps and large body movements). These results are shown in our accompanying video.

\subsection{Limitations}\label{sec:limitations}
In our method, two branches can help each other when at least one of the branches works well.
So, it is easy to imagine that if the two branches both face their inherent difficulties, our system will also suffer.
For example, when the subject moves out of the camera view for a long time, the visual branch has no input, and the drift of the IMU branch becomes large.
In such circumstances, IMUs are susceptible to error accumulation caused by magnetic disturbances.
We could consider modeling magnetic disturbance from combined modalities in the future.
%
Our method neglects the shape of different people and uses a mean shape for common adults.
Since we have visual signals, we could consider body shape in the future to get a more complete reconstruction.
\section{CONCLUSION}
In this paper, we present a method that fuses monocular images with 6 IMUs for real-time human motion capture, which can deal with extreme lighting, severe occlusion, and even subjects out of camera view.
The dual coordinate strategy makes the neural network better learn from the inertial signals in different cases.
The hidden state feedback enables the two branches in the dual coordinate strategy to exchange information from time to time, thus allowing the one with high confidence to help the other achieve better results. 
Quantitative and qualitative results show that our method can reconstruct more robust and accurate motion over state-of-the-art methods.

\begin{acks}
This work was supported by the National Key R\&D Program of China (2018YFA0704000), Beijing Natural Science Foundation (M22024), the NSFC (No.62021002), and the Key Research and Development Project of Tibet Autonomous Region (XZ202101ZY0019G).
This work was also supported by THUIBCS, Tsinghua University, and BLBCI, Beijing Municipal Education Commission.
Feng Xu is the corresponding author.
\end{acks}

\bibliographystyle{ACM-Reference-Format}
\bibliography{reference}
\end{document}